\documentclass{article}

\PassOptionsToPackage{numbers, compress}{natbib}

     \usepackage[final]{neurips_2023}

\usepackage[utf8]{inputenc} 
\usepackage[T1]{fontenc}    
\usepackage{hyperref}       
\usepackage{url}            
\usepackage{booktabs}       
\usepackage{amsfonts}       
\usepackage{nicefrac}       
\usepackage{microtype}      
\usepackage{xcolor}         
\usepackage{lipsum}
\usepackage{mwe}

\usepackage{microtype}
\usepackage{graphicx}
\usepackage{subfigure}
\usepackage{booktabs} 
 
\usepackage{hyperref}
\usepackage{amsmath}
\usepackage{amssymb}
\usepackage{mathtools}
\usepackage{amsthm} 
 
\graphicspath{ {./images/} }
\usepackage[utf8]{inputenc}

\newcommand{\Sb}{\mathbf{S}}
\newcommand{\Tb}{\mathbf{T}}

\newcommand{\Xb}{\mathbf{X}}
\newcommand{\Wb}{\mathbf{W}}
 
\newcommand{\ab}{\mathbf{a}}
\newcommand{\bb}{\mathbf{b}}
\newcommand{\fb}{\mathbf{f}}
\newcommand{\xb}{\mathbf{x}}

\theoremstyle{plain}

\theoremstyle{definition}

\theoremstyle{remark}

\usepackage[textsize=tiny]{todonotes}

\title{Iterated Piecewise Affine (IPA) Approximation for Language Modeling}

\author{%
  Davood Shamsi \\
  \texttt{davood@axiomsix.com} \\
   \And
   Wen-yu Hua \\
   \texttt{wenyu\_hua@apple.com} \\
   \AND
   Brian Williams \\
   \texttt{brian@vartia.ai} \\
}

\begin{document}
\maketitle

 \vspace{-0.2cm}
\begin{abstract}
In this work, we demonstrate the application of a first-order Taylor expansion to approximate a generic function $F: R^{n \times m} \to R^{n \times m}$ and utilize it in language modeling. To enhance the basic Taylor expansion, we introduce iteration and piecewise modeling, leading us to name the algorithm the Iterative Piecewise Affine (IPA) approximation. The final algorithm exhibits interesting resemblances to the Transformers decoder architecture. By comparing parameter arrangements in IPA and Transformers, we observe a strikingly similar performance, with IPA outperforming Transformers by 1.5\% in the next token prediction task with cross-entropy loss for smaller sequence lengths.
\end{abstract}
\vspace{-0.4cm}
\section{Introduction and Problem Description}
 
\begin{figure}
    \centering
    \begin{minipage}{0.45\textwidth}
        \centering
        \includegraphics[width=0.5\textwidth]{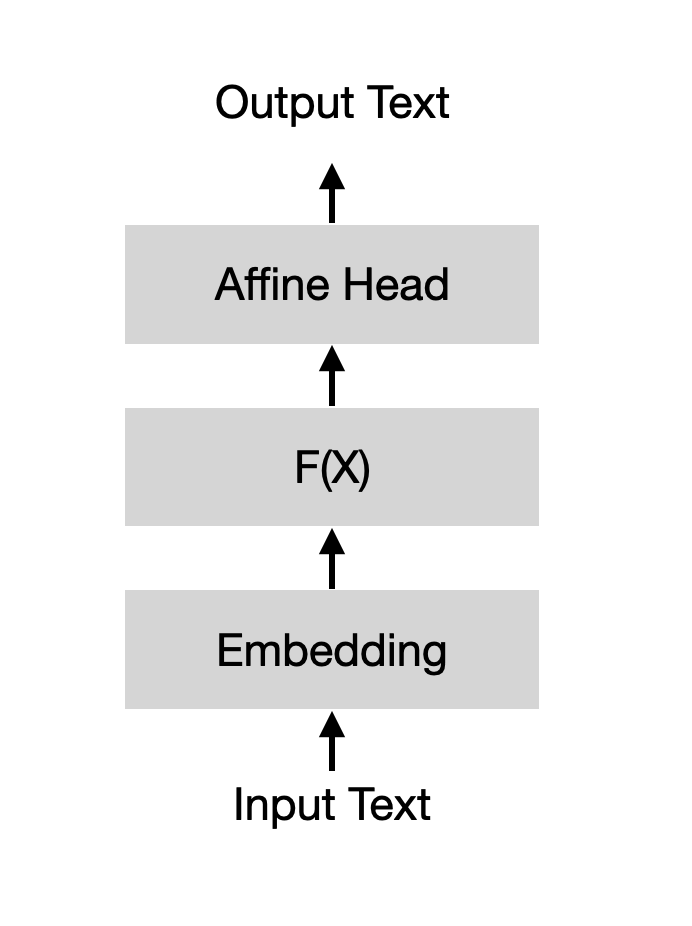} 
        \caption{Stages in language modeling: tokens are embedded into vectors, the resulting matrix is passed through a function $F(\Xb)$ and the next token is predicted with an affine head (typically a feedforward layer).}
        \label{fig:layers}
    \end{minipage}\hfill
    \begin{minipage}{0.45\textwidth}
        \centering
        \includegraphics[width=1.1\textwidth]{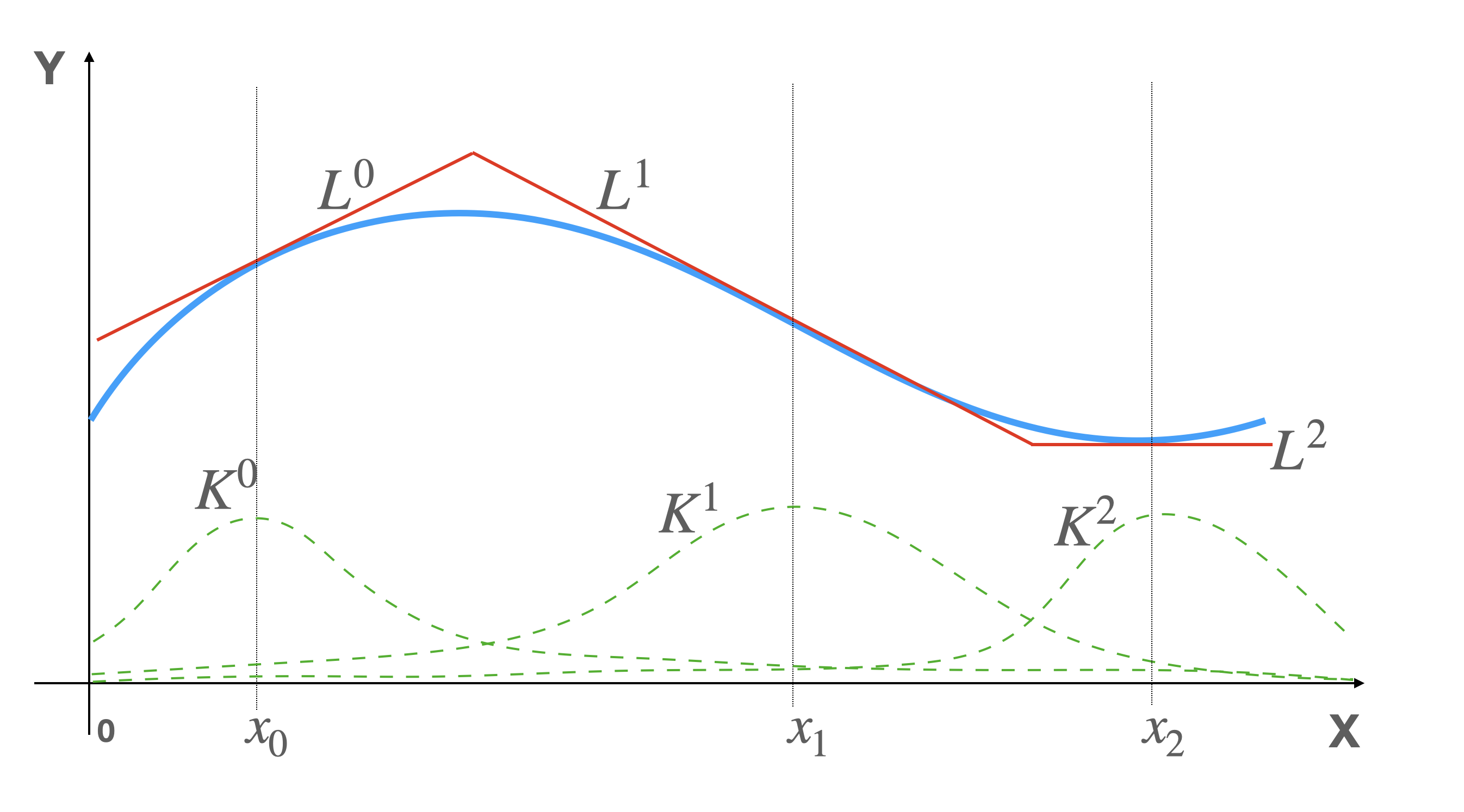} 
        \caption{Piecewise Affine Function Estimation: The blue curve is estimated using first-order Taylor expansions at points $x_0$, $x_1$, and $x_2$. The final estimate is a combination of these lines using kernel functions $K_0$, $K_1$, and $K_2$.}
        \label{fig:piecewise}
    \end{minipage}
\end{figure}

Transformers \cite{DBLP:journals/corr/VaswaniSPUJGKP17} and their variations \cite{DBLP:journals/corr/abs200914794, DBLP:journals/corr/abs200405150, DBLP:journals/corr/abs190409925, DBLP:journals/corr/abs190410509, DBLP:journals/corr/BritzGL17, DBLP:journals/corr/abs210609681, DBLP:books/lib/HastieTF09, DBLP:journals/corr/abs210314899, DBLP:conf/naacl/DevlinCLT19, DBLP:journals/corr/abs210700606, DBLP:journals/corr/abs201211747, DBLP:journals/corr/abs210414294, DBLP:journals/corr/abs211204446, DBLP:journals/corr/abs201009709, DBLP:journals/corr/abs200512766} have been the driver of the recent development in AI. However, the model architecture appears to be more the result of craftsmanship than formal function approximation methodology. In this paper, we demonstrate how a similar yet fundamentally different model can be developed by using Taylor expansion and piecewise function estimation techniques. 
 \cite{DBLP:journals/corr/abs200514165, DBLP:journals/corr/abs190711692, DBLP:conf/naacl/DevlinCLT19, DBLP:journals/corr/abs190908053, DD:PALM, DBLP:journals/corr/abs220108239}. 
In a language model \cite{DBLP:journals/corr/abs200514165, DBLP:journals/corr/abs190711692, DBLP:conf/naacl/DevlinCLT19, DBLP:journals/corr/abs190908053, DD:PALM, DBLP:journals/corr/abs220108239}, as shown in Fig. \ref{fig:layers}, first there is an embedding layer that maps each token to a vector in $R^n$. If the input sequence is of length $m$, the output of the embedding layer is a matrix $\Xb \in R^{n \times m}$. Next, the matrix $\Xb$ is passed through a function $F$. And finally, there is an affine head (with softmax) that maps the output of $F(\Xb)$ to probability distribution of the next word. While there may be an embedding layer and final prediction head, at its core, a language model approximates a function $F: R^{n \times m} \to R^{n \times m}$ that maps a matrix space to itself.
Once the language modeling task is mapped to the function approximation in the matrix domain, it is natural to ask how effective is a first-order Taylor expansion? Here is the Taylor expansion around a given point $x_0$ (in one dimension for simplicity):
\begin{equation}
       \label{eq:1d_taylor}
        F(x)  \approx L(x) = F(x_0) + F^\prime (x_0)(x - x_0).
\end{equation}
The first-order Taylor expansion can be a good approximation around the center point $x_0$, but not globally. To improve the accuracy of the approximation, we can write the first-order Taylor expansion around $P$ center points and combine them using a set of kernel functions:
\begin{equation}
                \label{eq:1d_piecewise}
                F(x) \approx \sum_{p=1}^PK^p(x)L^p(x).
                \vspace{-0.2cm}
\end{equation}
$K^p(x)$ and $L^p(x)$ are kernel functions (e.g. exponential) and affine approximations Eq. (\ref{eq:1d_taylor}) for $p$-th center point.
For visual illustration, the three red lines in Fig. \ref{fig:piecewise} are Taylor expansions around 3 center points and we used kernels (dashed-green line) to combine them. Upon closer inspection, we can observe that estimating through multiple center points exhibits a very similar, though not identical, relationship to the multi-head architecture found in Transformers. Finally, we apply this approximation iteratively by creating layers to arrive at the final estimator. The parameters of the Tylor expansions are calculated using gradient descent on a training dataset. 
Our main contribution is the introduction of the Iterative Piecewise Affine (IPA) algorithm, which is a straightforward and mathematically intuitive method for approximating a function in the matrix space.  The IPA algorithm is competitive to Transformers, but it does not use any heuristics and is easy to understand. 

This paper is organized as follows. First, in Section \ref{sec:ipa_main}, we extend Eq. (\ref{eq:1d_taylor}) and (\ref{eq:1d_piecewise}) from one-dimensional functions to functions in the matrix domain. Then, in Section \ref{sec:ipa4llm}, we modify the basic IPA to make it suitable for language modeling (e.g., causality constraint). Next, in Section \ref{sec:relationship_w_transformers}, we demonstrate the relationship between IPA and Transformers. Finally, in Section \ref{sec:results}, we compare the performance of IPA to Transformers and conclude in Section \ref{sec:conclusion}.

\vspace{-0.15cm}
 
\section{Iterative Piecewise Affine Estimator (IPA) Approximation}
\vspace{-0.15cm}

\label{sec:ipa_main}
Our goal is to estimate a function $F: R^{n \times m} \to R^{n \times m}$ that maps a matrix space. We use affine function estimators based on first-order Taylor expansion, piecewise affine estimation using kernel functions, and improve the estimator through iteration.
 
\subsection{Affine Estimation}
\vspace{-0.1cm}

\label{sec:affine}
Lets consider writing Taylor expansion for the rows and columns of the function $F$ separately.
 In what follows, as shown in Fig. \ref{fig:matrix}, we use a dot before the index of a variable to denote that it represents a column and a dot after the index to show that it represents a row.


\textbf{Column Representation}       
Let $\fb_{.j}(\Xb)$ be the $j$-th column of $F(\Xb)$ and $\hat{\fb}_{.j} (\Xb)$ be its first-order Taylor approximation. Then,
\begin{equation}
        \label{eq:taylor_column_matrix}
        \fb_{.j}(\Xb)  \approx \hat{\fb}_{.j}(\Xb) = \ab_{j} + \sum_{l=1}^m \Sb_{j,l}  \xb_{.l}, 
         \vspace{-0.2cm}
\end{equation}
where, $\ab_j \in R^n$ and $\Sb_{j,l} \in R^{n \times n}$ are the approximation coefficients and $\xb_{.l}$ is the $l$-th column of $\Xb$.  
       
\textbf{Row Representation}
Similarly, we can write the Taylor expansion for rows. If $\fb_{i.}(\Xb)$ is the $i$-th row of $F(\Xb)$, and $\xb_{r.}$ is the $r$-th row of $\Xb$, then 
\begin{equation}
        \label{eq:taylor_row_matrix}
        \fb_{i.}(\Xb) \approx \hat{\fb}_{i.}(\Xb) = \bb_{i} + \sum_{r=1}^n \Tb_{i,r}  \xb_{r.},
         \vspace{-0.2cm}
\end{equation}
where, $\bb_i \in R^{m}$ and $\Tb_{i,r} \in R^{m\times m}$ are the row approximation coefficients.

\subsection{Piecewise Affine Estimation}         
\vspace{-0.1cm}                                                                                                                                                             
\label{sec:piecewise}
 
In the previous section, we used first-order Taylor expansion to estimate $F$. However, Taylor expansion around one point might not be an accurate estimator over the general domain of the function. To address this issue, we can write the expansion around multiple points and combine them using some kernel functions  (e.g. see \cite{DBLP:books/lib/HastieTF09} page 172).  
Eq. (\ref{eq:taylor_column_matrix}),   $\Sb_{j,l}$  can be estimated as:
\begin{equation}
	\label{eq:piecewise}
       \Sb_{j,l}(\Xb) \approx \sum_{p=1}^P K_{j,l}^p(\Xb) \Sb^p_{j,l},
\end{equation}
where $\Sb^p_{j,l}$ are coefficients for $p$-th Taylor expansions. We will discuss the choice of kernels $K_{j,l}^p(\Xb) $ later in Section \ref{sec:kernel}.
Please note the concept of kernel used here is different from the one used in Transformers, e.g. \cite{DBLP:journals/corr/abs200914794, DBLP:journals/corr/abs200405150}. The same estimation can be applied to $\Tb_{i,r}$ in Eq. (\ref{eq:taylor_row_matrix}),
 
\subsection{Estimating through Iterated Functions}     
\vspace{-0.05cm}                                                                                                                                                           
\label{sec:nested}
As it is common in the deep learning literature \cite{DBLP:journals/corr/HeZRS15,DBLP:journals/corr/HuangALS17}, we can use iteration to model higher levels of non-linearities: 
\begin{equation}
                \label{eq:expansion}
                F(\Xb) \approx \hat{F}_{\gamma_1} \circ \hat{F}_{\gamma_2} \circ \hat{F}_{\gamma_3} \circ \dots \circ \hat{F}_{\gamma_n}(\Xb),
\end{equation}
where symbol $\circ$ represents ``function of function'' and $\hat{F}_{\gamma_i}$ are, alternatively, piecewise affine estimations from Eq. (\ref{eq:taylor_column_matrix}) and (\ref{eq:taylor_row_matrix}) with different estimation parameters. 

\begin{figure}
    \centering
    \begin{minipage}{0.45\textwidth}
        \centering
\small
\(
\Xb= 
\begin{pmatrix}
 \xb_{.1} &
 \xb_{.2} & 
 \dots &
 \xb_{.m} \\
\end{pmatrix}
,
\Xb= 
\begin{pmatrix}
 \xb_{1.} \\
 \xb_{2.} \\
 \vdots \\
 \xb_{n.} \\
\end{pmatrix}
\)
\(
F(\Xb)= 
\begin{pmatrix}
 \fb_{.1}(\Xb) &
 \fb_{.2}(\Xb) & 
 \dots &
 \fb_{.m}(\Xb) \\
\end{pmatrix}
, F(\Xb)= 
\begin{pmatrix}
 \fb_{1.}(\Xb) \\
 \fb_{2.}(\Xb) \\
 \vdots \\
 \fb_{n.}(\Xb) \\
\end{pmatrix}
\)
\caption{Column and row representation of $\Xb$ and $F(\Xb)$. These representations can result in different approximations.}
\label{fig:matrix}
        
    \end{minipage}\hfill
    \begin{minipage}{0.43\textwidth}
        \centering
        \includegraphics[width=0.95\textwidth]{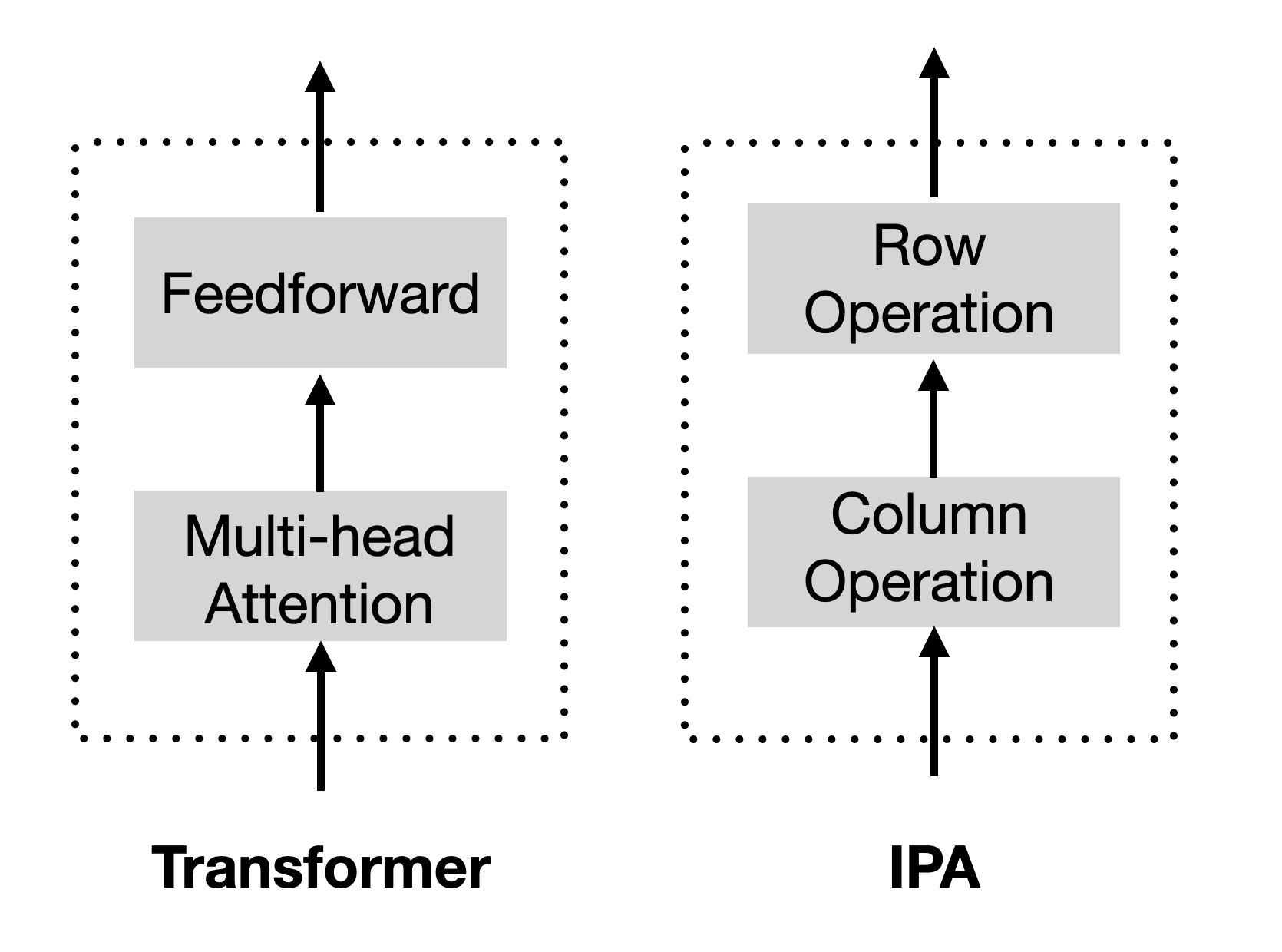}
        \caption{Comparing the IPA algorithm to Transformers. Both methods apply column and row operations consecutively.}
        \label{fig:ipa_transformer}
    \end{minipage}
\end{figure}

 \vspace{-0.15cm}
\section{IPA Approximation for Language Modeling}
\vspace{-0.15cm}
\label{sec:ipa4llm}
As is commonly found in literature, we formulate language modeling by estimating a shifted sequence of tokens from its original form. Fig. \ref{fig:layers} illustrates the end-to-end language modeling task, with the main objective being to approximate $F(\Xb)$. The input text is first tokenized and embedded into vectors, which are then fed into the IPA to approximate $F(\Xb)$. Finally, a commonly-used affine head predicts the output (next tokens). We need to make some modifications to the original IPA formulation to make it compatible with language modeling, which will be discussed below.
\subsection{Adding Causality Constraints to the IPA Algorithm}   
\vspace{-0.05cm}
To ensure that we only utilize past tokens for predicting the next token in both the affine functions and kernels, we set all coefficients from future tokens to zero. 
 
\textbf{Column Operation}: To mask the future tokens, in Eq. (\ref{eq:taylor_column_matrix}), the sum should be from $l=1$ to $j$. 


\textbf{Row Operation}:         
Enforcing the causality constraint for row approximation can be more challenging. For simplicity, we make the constraint stronger by enforcing the matrix $\Tb_{i,r}$ in Eq. (\ref{eq:taylor_row_matrix}) to be diagonal.

\subsection{Kernel Function}
\vspace{-0.05cm}
\label{sec:kernel}
As has been proven effective in natural language processing literature, we use attention-style kernel functions for column representation: $K_{j,l}^p(\Xb) = \lambda^{-1} e^{\xb_{.l}^T \Wb^p \xb_{.j} },$
where $\Wb^p \in R^{n \times n}$ are parameters of the kernel, and $\lambda$ is the normalization factor, and chosen such that $ \sum_{p} K_{j,l}^p(\Xb) = 1$. For the row representation, we use Gaussian radial kernels (Section 5.8.2 in \cite{DBLP:books/lib/HastieTF09}). 
      
    
\subsection{Position Independent Mapping}
\vspace{-0.05cm}
In language modeling, each column of matrix $X$ represents a token in the input sequence. Therefore, it is natural to assume that the position of the token should not affect the mapping:
$\Sb_{j,l} = \Sb_{j^\prime,l} \;\;\; \forall j, j^\prime,$; $\Tb_{i,r} = \Tb_{i^\prime,r} \;\;\; \forall i, i^\prime.$ This assumption is mainly for reducing number of parameters, and position of the token is still important in the IPA formulation.

\subsection{Reducing Parameters with Low Rank Matrices}
\vspace{-0.05cm}
\label{sec:low_rank}
In the previous formulation, there were no restrictions on $\Sb_{i,l}$ and $\Wb^p$, so they could be full-rank matrices. To lower the number of parameters, we can assume that they have a lower rank of $k$.
For example: $\Wb^p = \Wb^p_l \times \Wb^p_r$, where $\Wb^p_l \in R^{n\times k}$ and $\Wb^p_l \in R^{k\times n}$.

 \vspace{-0.15cm}
\section{Relationship with Transformers}
\vspace{-0.15cm}
\label{sec:relationship_w_transformers}
 
Although there are fundamental differences between the IPA algorithm and Transformers, as illustrated in Fig. \ref{fig:ipa_transformer}, their architectures share some intriguing similarities. Specifically, the multi-head attention mechanism in Transformers can be viewed as a column operation and the subsequent feedforward layer as a row operation. Additionally, upon closer analysis, it can be observed that the kernels in the piecewise affine operation of the IPA algorithm have similar roles as attention heads in Transformers. 
 

\vspace{-0.15cm}
\section{Experimental Results}
\vspace{-0.15cm}
\label{sec:results}
 
 
 

In this section, we compare performance of the IPA algorithm with GPT architecture \cite{DD:RKT2018} (stack of Transformers decoder) on the WikiText103 dataset \cite{DBLP:journals/corr/MerityXBS16}. In order to make a meaningful comparison, we closely matched the internal parameters of IPA and Transformer. For all experiments, the embedding size was set to $n=120$ and there were 4 layers. For the column operation, we set the number of affine functions equal to the number of heads in the Transformer model ($P_{\mbox{column}}=8$) and the rank of matrices ($k$ in Section \ref{sec:low_rank}) equal to the embedding size of each head in the Transformer model ($k=15$). For the row operation, we set the number of affine functions equal to the ratio of the feedforward's inner dimension to the embedding size. Specifically, we set Transformer feedforward's inner dimension equal to 4 and thus, $P_{\mbox{row}}=4$. We use Byte Pair Encoding \cite{DBLP:conf/dcc/Bloom96, DBLP:journals/corr/SennrichHB15} to tokenize the input text, and no dropout was used for either model.
  \vspace{-0.4cm}
\begin{table}[!h]
\caption{Train and test loss on WikiText103 dataset. Loss is the cross-Entropy for the next word prediction, and $m$ is the sequence length. Time per iteration is measured in milliseconds.}
\label{tbl:results}
\begin{center}
\begin{tabular}{ |c|c|c|c|c| } 
 \hline
 Model, & \# Parameters & Train  & Test  & Time per  \\ 
($m$)  & &  Loss &  Loss & Iteration\\ 
 \hline\hline
  GPT, (100) & 4.45M & 4.51 & 4.45 & 28.4\\ 
\hline 
 IPA,  (100)  & 4.49M & 4.49 & 4.38 & 30.7\\ 
\hline 
GPT,  (250) & 4.47M & 4.13 & 4.09 & 65.0\\ 
\hline 
 IPA, (250)  & 4.56M & 4.17 & 4.07 &66.5\\ 
 \hline
  GPT, (500) & 4.50M & 3.91 & 3.90 & 148.8\\ 
\hline 
 IPA, (500)  & 4.68M & 3.98 & 3.89 & 147.0\\ 
\hline 
\end{tabular}
\end{center}
\end{table}

Table \ref{tbl:results} displays the train and test loss of the IPA algorithm compared to the GPT architecture (Transformer decoders) for three different sequence lengths on the WikiText103 dataset. As reminder, variable $m$ represent length of the sequence. The loss is calculated as cross-entropy for the next token prediction. All experiments were run until convergence based on the test loss ($\approx 10$ million steps with a learning rate of 2e-5). As shown in Table \ref{tbl:results}, with a similar configuration, the IPA algorithm has better performance than GPT for small sequence lengths ($1.5\%$ for $m=100$) but they have very similar performance for longer sequences ($m=500$). From Table \ref{tbl:results}, you can observe that the training time ($\approx$ computation cost) is very similar in both models.

 
 
 \vspace{-0.15cm}
\section{Conclusion}
\vspace{-0.15cm}
\label{sec:conclusion}
In this paper, we introduced IPA algorithm for estimating a general function $F: R^{n \times m} \to R^{n \times m}$ and applied it to language modeling. The IPA algorithm is straightforward, intuitive, and shows comparable performance to Transformers.

\newpage
\bibliographystyle{achemso}
\bibliography{ipa_neurips_2023.bib}

\end{document}